\documentclass[11pt]{article}

\usepackage[final]{acl}

\usepackage{times}
\usepackage{latexsym}
\usepackage[T1]{fontenc}

\usepackage[utf8]{inputenc}
\usepackage{microtype}

\usepackage{inconsolata}
\usepackage{booktabs}
\usepackage{graphicx}
\usepackage{amsmath, amssymb}
\usepackage{float}

\usepackage{natbib}
\usepackage{algorithm}
\usepackage{algpseudocode}
\usepackage{subcaption}
\usepackage{enumitem}

\title{Entropy-aware Masking for Masked Language Modeling}

\author{Gokul Srinivasagan \and Kai Hartung \and Munir Georges \\
  AImotion Bavaria, Technische Hochschule Ingolstadt, Germany\\
  \texttt{{\{Gokul.Srinivasagan, Kai.Hartung, Munir.Georges\}@thi.de}} }

\begin{document}
\maketitle
\begin{abstract}
Masked language modeling has become a standard pretraining objective for training encoder-based language models. In this approach, certain tokens in the input are masked, and the model learns to predict them using the surrounding context. This process enables the model to capture both syntactic and semantic properties of language. Conventionally, the tokens selected for masking are chosen at random, which may not always yield the most effective learning signals. In this work, we examine a token masking strategy based on entropy distribution. We use the model’s entropy over token predictions to identify which tokens should be masked. This method aims to target tokens that are more informative and uncertain to improve the training efficacy. We also propose a novel self-masking approach that enhances training efficiency without relying on an external reference model. Experimental results demonstrate that our method achieves an average performance improvement of 5\% in GLUE scores compared to the baseline. Further, we experiment with combining knowledge distillation with entropy masking, resulting in the best overall results. 
\end{abstract}

\section{Introduction}
\label{sec:intro}

The last few years have witnessed groundbreaking advancements in the field of natural language understanding, largely driven by the emergence of large pretrained language models \citep{zhao2023survey}.
Their success is primarily attributed to a two-stage training process \citep{devlin-etal-2019-bert, liu2019roberta}: an initial self-supervised pretraining phase, followed by supervised fine-tuning for specific downstream tasks like sentiment classification. 
For encoder-based architectures, such as BERT \citep{devlin-etal-2019-bert}, the pretraining objective typically involves masked language modeling (MLM). 
This training paradigm involves replacing tokens in the input sequence with mask tokens and training the model to predict these masked tokens using the surrounding context.
This enables the model to learn both syntactic structure and semantic understanding of language \citep{rogers2021primer}. 
However, these MLM approaches rely on randomly selecting tokens to be masked, which may not be the most effective strategy for learning meaningful representations \citep{gu-etal-2020-train, lad2022using}.

So in this work, we propose an alternative method for the selection of masked tokens during pretraining by utilizing entropy.
In the context of language models, entropy measures the uncertainty in the model’s prediction over the vocabulary. 
We can utilize this entropy to identify tokens and contexts in which the model is uncertain in its prediction and select tokens to mask, specifically to address this uncertainty.
We explore with two sources for the entropy:
(1) entropy computed from an external pretrained reference model, and (2) our novel self-masking approach, where entropy is computed from the target model itself during training.
To fully utilize the reference model, we also explore the combination of our approach with knowledge distillation, since this also uses a teacher model as reference for training the student model. We also perform several ablation studies like increasing the size of the pretraining dataset, varying the masking percentage and fine-tuning with weight freezing to test the effectiveness of our proposed approach.

\section{Method} \label{sec:method}

\begin{figure}[!ht]
    \centering
    \includegraphics[width=0.45\textwidth]{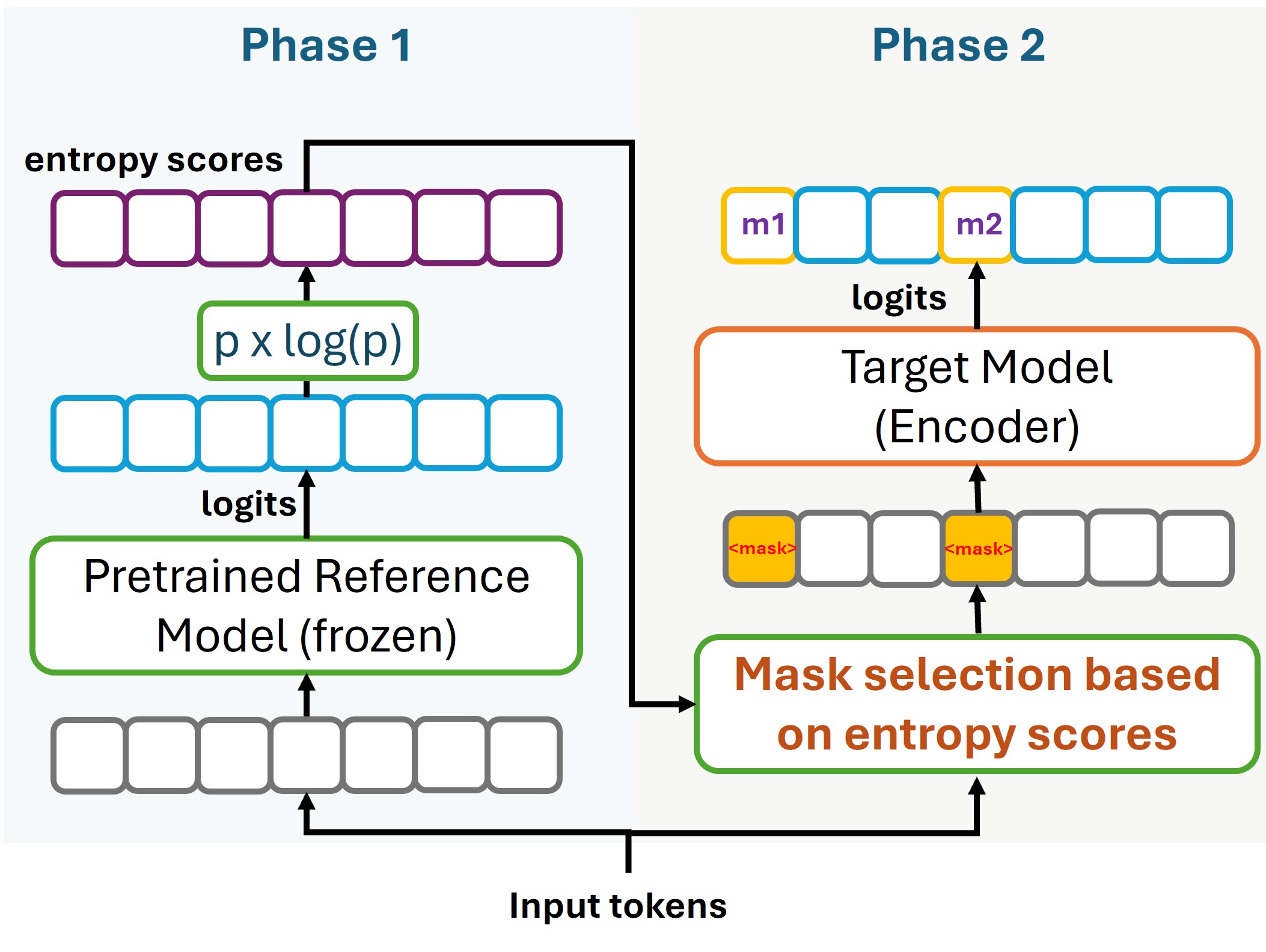}
    \caption{
        The overall process of entropy-aware masked language modeling using teacher-masking.
    }
    \label{fig:method}
\end{figure}

First, we define the training objective for masked language modeling.
$\mathcal{V}$ is the vocabulary available to the language model $P_{\theta}$ with tokens $t \in \mathcal{V}$.
$\mathbf{T} = t_1...t_n$ is the ground truth sequence of $n$ tokens.

For the masking, we first define the mask $\mathcal{M} \subset \{1, 2, ..., n\}$ as a subset of indices and $i \in \mathcal{M}$ as indices of tokens $t_i$ to be masked.
As a replacement for these tokens, we add a special mask token $t_m$ to the vocabulary.
This mask is applied to produce a masked sequence $\mathbf{T_\mathcal{M}} = t'_1...t'_n$ for $\mathbf{T}$, such that 
\begin{equation*}
    t'_j = \begin{cases}
        t_m &\text{if } j \in M \\
        t_j &\text{else}
    \end{cases}
\end{equation*}

The masked sequence $\mathbf{T}_{\mathcal{M}}$ serves as input to the model $P_{\theta}$ with parameters $\theta$. 
It is tasked with predicting the original tokens $t_i$ for each mask token $t_m$.
The respective loss is computed as the negative log likelihood the model assigns to the replaced tokens $t_i$:
\begin{equation*}
    \mathcal{L}_{\text{MLM}} = - \sum_{i \in \mathcal{M}} \log P_\theta \left( t_i \mid \mathbf{T}_{\mathcal{M}} \right)
\end{equation*}

Our baseline model uses random masking for training, so each position in $\mathbf{T}$ is a possible candidate for masking the token at that position with an equal probability $p$.

\subsection{Entropy masking}

Instead of random selection, we now introduce entropy-based selection, visualized in figure \ref{fig:method}.
Entropy measures a model's uncertainty in its prediction.
When the entropy for a token is high, the model's probability distribution over possible choices from the vocabulary is nearly uniform.
This indicates uncertainty since many tokens are similarly likely to fit. 
Conversely, when the entropy is low, the distribution is sharply peaked, so only a few tokens have a high probability to fit in the context, and the model's confidence in these tokens is high.
With this adaptation, we intend to focus the model's learning on those tokens about which the model is uncertain and, in turn, include fewer repetitions on tokens the model is already certain about.
With this focus, the feedback for any single prediction is expected to be more informative.

We access this uncertainty in form of the entropy over the model's probability assignments for each token in a given sequence.
The entropy $e_{t_j}$ for a token $t_j$ is computed from the model's probability estimates for each option $v \in V$ as:
\begin{equation*}
    e_{t_j} = - \sum_{v \in V} P_{t_j}(v)\log P_{t_j}(v)
\end{equation*}

The estimate for the probability distribution over $V$ at a position $i$ in the sequence is in turn computed from the model's output logits $f_{\theta}(\mathbf{T})$ for the input sequence $\mathbf{T}$:
\begin{equation*}
    \mathbf{P}_{t_j}(V) = softmax(f_{\theta}(\mathbf{T})_j)
\end{equation*}

\subsection{Entropy sources} \label{sec:method:sources}

We derive the entropy in two different variants.

\textbf{Teacher-masking} is the use of a pretrained external reference model to estimate the token entropy.
Similar to knowledge distillation, the pretrained model guides the model to be trained, even though no predictions by the teacher are directly available to the student model in our case.
As this external teacher model does not change during the training process, the selection of the masks is necessarily static and the learning signals may become less informative as training goes on.

\textbf{Self-masking}: To reduce the influence of the external reference model, we use the target model itself to estimate the entropy.
This allows us to update the masks as the model learns and to align them with what the model currently is least certain about. 
Our self-masking has a problem at the start of the training, as the uninitiated model does not yet provide any informative probability estimates.
To mitigate this cold start issue, we start the training with an initial teacher-masking phase, before switching to the self-masking.

\subsection{Selection strategies} \label{sec:method:selection}

We explore four different strategies for the usage of the token-wise entropy $e_t$ to select mask $\mathcal{M}$.
For each strategy, we select $k$ tokens to be masked, where $k$ is 15\% of the current sequence length.

\textbf{High entropy masking}. 
This strategy selects the $k$ tokens with the highest entropy in the sequence.
This means, in a context of comparatively well understood tokens, the model has to predict the tokens with the highest uncertainty.

\textbf{Low entropy masking} selects the $k$ tokens with the lowest entropy in the sequence.
In this variant, the model has to predict the best understood tokens from a comparatively uncertain context.

\textbf{Mid-entropy masking}. 
In this strategy, the mid-range entropy values are selected to be masked. This is a control strategy to test that the extreme values of the entropy are relevant.

\textbf{Marginal entropy masking}.
In this strategy, which is the inverse of the mid-entropy strategy, half of the masked tokens are in the high entropy range and half in the low entropy range, so that mid-entropy tokens are not masked.

\textbf{Alternating entropy masking}. 
The tokens are masked alternatively over each batch.
With a probability of 50\%, a batch is masked using either high or low entropy masking.
The model has to predict uncertain tokens in a certain context and vice versa.

\subsection{Implementation Details}
\label{sec:implement}

We train two variations of BERT models with varying configurations as depicted in the table below:

\begin{table}[H]
    \centering
    \begin{tabular}{l|llll}
     \hline
        \textbf{Model}   & \textbf{\#Layer} & \textbf{Dim.} & \textbf{\#Attn} & \textbf{\#Param}\\ \hline
        \textbf{BERT}   & 12    & 768  & 12 &  109M        \\
        \textbf{BERTlet}   & 4     & 512  & 8 & 28.8M \\ \hline       
    \end{tabular}
\end{table}

We use only the MLM objective for our experiments. As reference models, we use pretrained BERT models of the same dimensions that are available on Hugging Face \cite{wolf-etal-2020-transformers}. We pretrain the models on the wikitext-103 dataset \cite{merity2016pointer}, which contains around 30k samples from Wikipedia articles. We utilize 95\% of the data for training and reserve 5\% for testing. Each model is pretrained for 25 epochs with a learning rate of 1e-4. 
Unless specified, all our models use a 15\% masking ratio for masked language modeling.
During fine-tuning, the models are trained for 50 epochs with early-stopping with patience value 5 and a learning rate of 5e-5. 
The models are trained with the mixed precision format BF16. 
For fine-tuning we use a batch size of 256. 
All our experiments are carried out on a single Nvidia A100 80GB GPU. 
For training, we use the transformers \cite{wolf-etal-2020-transformers} and datasets \cite{lhoest-etal-2021-datasets} libraries.
We run all our experiments twice with different seed values and report the mean values as results. 

\section{Experiments} \label{sec:eval}

\subsection{Teacher-masking}

\begin{table*}[!ht]
    \centering
    \begin{tabular}{l|lllllll|l}
    \toprule
        \textbf{Model} & \textbf{CoLa} & \textbf{MRPC} & \textbf{QNLI} & \textbf{QQP} & \textbf{RTE} & \textbf{SST-2} & \textbf{MNLI} & \textbf{Total } \\ \toprule
        \textbf{BERT} (baseline) & 69.42 & 81.80 & 74.98 & 83.75 & 52.35 & 82.34 & 68.25 & 73.27  \\ 
        \textbf{BERTlet} (baseline) & 69.32 & 81.79 & 65.20 & 82.24 & 53.07 & 79.36 & 65.51 & 70.93  \\ \midrule
        
        \textbf{BERT$_{max}$} & 67.98 & 78.05 & 80.72 & 86.15 & 52.71 & 85.78 & 73.76 & \textbf{75.02 } \\ 
        \textbf{BERTlet$_{max}$} & 68.74 & 81.01 & 76.04 & 84.67 & 56.32 & 81.54 & 70.19 & \textbf{74.07 } \\ \midrule
        
        \textbf{BERT$_{min}$} & 69.42 & 81.03 & 79.68 & 85.01 & 54.51 & 81.31 & 72.65 & \textbf{74.80} \\ 
        \textbf{BERTlet$_{min}$} & 69.03 & 79.68 & 65.62 & 82.60 & 51.62 & 80.62 & 69.91 & \textbf{71.30}  \\ \midrule
        
        \textbf{BERT$_{marg}$} & 69.51 & 79.53 & 82.13 & 86.38 & 54.51 & 83.14 & 72.58 & \textbf{75.40}  \\
         \textbf{BERTlet$_{marg}$} & 69.03 & 79.94 & 66.14 & 82.42 & 54.51 & 80.39 & 69.93 & 64.17  \\ \midrule
         
        \textbf{BERT$_{alt}$} & 68.74 & 81.07 & 82.92 & 86.71 & 53.07 & 86.58 & 73.99 & \textbf{76.15}  \\ 
        \textbf{BERTlet$_{alt}$} & 68.84 & 81.37 & 65.95 & 82.20 & 53.43 & 80.96 & 70.26 & 62.79  \\ \midrule

        \textbf{BERT$_{mid}$} & 69.13 & 81.22 & 50.54 & 63.18 & 52.71 & 50.92 & 35.22 & 57.56  \\
        \textbf{BERTlet$_{mid}$} & 69.13 & 81.22 & 60.66 & 80.74 & 52.71 & 79.82 & 59.85 & 62.23  \\  \bottomrule
    \end{tabular}
    \caption{
        Performance of models trained with teacher-masking on the GLUE benchmark. 
        Total is the average score over all tasks.
        $max$, $min$ and $mid$ describe high, low and mid-entropy token masking respectively, $marg$ masks half high and half low entropy tokens and $alt$ masks randomly either high or low entropy tokens.
        }
    \label{tab:result_ent}
\end{table*}

Table \ref{tab:result_ent} shows the performance of the BERT and BERTlet models trained with the teacher-masking approach. 
The model subscripts indicate the mask selection strategy used, as described in section \ref{sec:method:selection}.
The results show that high entropy masking produces improvements of 2 to 3 points for both BERT and BERTlet models.
This shows that the model learns semantic relationships between comparatively less surprising tokens and highly unpredictable tokens.
The low entropy strategy also improves, but not as much as the high entropy masking.
The BERTlet model in particular benefits from the high entropy version.
The marginal and alternating entropy strategies help the BERT model but hurt the smaller BERTlet model.
These show that the influence depends on model size and that extreme entropy masking provides more reliable improvement than the mixed approaches. 
Using the mid-entropy strategy results in a strong deterioration for both model sizes and for the larger BERT in particular. 
This further confirms that it is meaningful to use the extreme entropy values. All further experiments are derived from the most successful constellations of these experiments.

\subsection{Self-masking}

\begin{table*}[!ht]
    \centering
    \begin{tabular}{l|lllllll|l}
    \toprule
        \textbf{Model} & \textbf{CoLa} & \textbf{MRPC} & \textbf{QNLI} & \textbf{QQP} & \textbf{RTE} & \textbf{SST-2} & \textbf{MNLI} & \textbf{Total } \\ \toprule
        
        \textbf{BERT$_{max,cold}$} & 69.13 & 81.22 & 49.46 & 63.18 & 52.71 & 50.92 & 35.22 & 57.41 \\ 
        \textbf{BERTlet$_{max,cold}$} & 69.13 & 80.30 & 60.77 & 81.02 & 54.51 & 77.87 & 62.00 & 61.37 \\ \midrule
        
        \textbf{BERT$_{max,init}$} & 69.42 & 81.22 & 84.81 & 89.04 & 52.35 & 86.81 & 71.62 & \textbf{76.47 } \\ 
        \textbf{BERTlet$_{max,init}$} & 68.84 & 80.39 & 78.82 & 85.80 & 55.96 & 83.72 & 72.54 & \textbf{75.15} \\ \bottomrule
    \end{tabular}
    \caption{
        Performance of models trained with self-masking on the GLUE benchmark. 
        Total is the average score over all tasks.
        $max$ marks the high entropy token masking approach, $cold$ models are trained with self-masking only, while $init$ refers to initial teacher-masking followed by self-masking.
        }
    \label{tab:result_ent_self}
\end{table*}

Building on the teacher-masking results, we use high entropy masking as the best selection strategy to explore self-masking training.
The cold-start models purely trained with self-masking are denoted as 'cold' in table \ref{tab:result_ent_self}, while the models initially trained using the teacher-masking are subscripted with 'init'.
As expected, the pure self-masking training does suffer from the cold start, resulting in the deteriorated performance for the resulting models.
In contrast, the reference-initialized models improve over the baseline by 4 points and 3 points for the BERTlet and BERT models respectively.
They improve by 1 point each over the teacher-masked high-entropy models, indicating a positive effect of the self-masking after initialization.

\subsection{Fine-tuning with weight freezing}

The language models can learn in both pretraining and fine-tuning phases of the training. 
To ensure the improvement observed in the downstream tasks doesn't come mainly at the later stage of training, we freeze the BERTlet model pretrained with high entropy masking and only train the classification layer for one epoch. 
The results reveal an improvement of 3\% over the baseline proving the effectiveness in the learning process during pretraining.

\subsection{Effect of masking percentage}

To explore the effect of a different percentage of masked tokens \cite{wettig-etal-2023-mask}, we also run high entropy masking experiments with BERT$_{max}$ and BERT$_{min}$.
On one hand we increase the masking percentage to 40\% and on the other we reduce it to masking only one token per sequence.
For the increased masking percentage, we observe no significant improvement over the default 15\%.
For the single token masking, we observe a drop of around 1\% in GLUE score. 
Based on these results, we find that a varying masking percentage has no to small negative impact.

\subsection{Effect of a larger training corpus}

To assess the impact of a larger pretraining dataset on performance, we perform experiments with the bookcorpus dataset \citep{Zhu_2015_ICCV}, which contains around 4.9B samples. 
We train a BERTlet$_{max,init}$ model on this dataset and observe an average GLUE score of 77.09, which is 2 points higher than the results from the wikitext-103 dataset and 5 points higher than the baseline BERTlet trained on bookcorpus (71.66).
This highlights the influence of a larger dataset in capturing complex linguistic patterns as it improves the performance of the downstream tasks.

\subsection{Effect of knowledge distillation}

As we are already working with a reference model to achieve the best improvements with our approach BERTlet$_{max,init}$, we explore if we can get further improvement by using the teacher model to guide the student model from both ends of training.
On one hand, knowledge distillation guides the student's output distribution by computing an additional loss from the difference between student and teacher prediction.
On the other hand, the initial input masking guides the student's focus in training by masking tokens based on the entropy.

All our knowledge distillation experiments give equal weights to loss from gold labels and teacher labels. 
We use BERTlet$_{max,init}$ self-masking models with two approaches to model initialization: i) Complete transfer and ii) Transfer after initialization. 
The complete transfer approach uses knowledge distillation from the start of the training.
The second approach uses only teacher-masking in the initial epoch and the knowledge distillation sets in only in the following epochs together with the self-masking. 
This provides some additional freedom for the target model in contrast to the complete transfer approach.
Our experimental results show that the model trained with transfer after initialization (77.19\%) outperforms the complete transfer approach (76.04\%) by around 1\% point GLUE score. 
The slightly worse performance may be due to an overfitting of the model due to the forced alignment by both entropy masking and distillation in the initial phase. But both of these two approaches outperform the model trained only with knowledge distillation.

\subsection{Weight analysis}

To identify which architectural components are most impacted during fine-tuning, we analyze how the weights of the fine-tuned models diverge from the pretrained state for BERTlet$_{max,init}$.
We extract the weight values for the pretrained model and the fine-tuned models for each of the GLUE tasks and compute the difference for each fine-tuned to the pretrained models. 
The feed-forward network in the final layer consistently exhibits the highest degree of deviation across all tasks. 
Among the fine-tuned models, the one trained on MNLI displays the greatest deviation at 7077.23\%, while the model fine-tuned on CoLA shows the smallest variation at 167.33\%. 
These results suggest that the size of the fine-tuning dataset significantly influences the degree of model variability.

\section{Conclusion}
In this work, we explored several entropy token selection strategies.
When using an external reference model, high entropy token selection gives the best performance, followed by low entropy selection.
Self token masking with weight initialization on initial iterations helps to avoid the cold start problem and offers the best results.
Even on its own, the self-masking approach proves to offer comparative performance to knowledge distillation.
These results can be further improved upon by combining the approach with knowledge distillation.

\section*{Limitations}
Due to computational limitations, we could not experiment with models with parameters more than 110 million. 
All our models are pretrained on English datasets. 
We have not experimented with other multilingual datasets, leaving room for future work. We also plan to investigate the effectiveness of our approach on other model architectures as part of future work \cite{sun-etal-2020-mobilebert, Lan2020ALBERT}.

\section*{Acknowledgements}

We thank the anonymous reviewers for their constructive feedback on our paper. This work has been supported by the Verkehrsverbund Großraum Ingolstadt (VGI) as part of the project newMIND.

\bibliography{custom}

\end{document}